\documentclass[a4paper,conference]{IEEEtran}

\usepackage{amsmath}
\usepackage{amssymb}
\usepackage{bm}
\usepackage{hyperref}
\usepackage{graphicx}
\usepackage{booktabs}
\usepackage{multirow}
\usepackage{tabularx}

\def\onedot{.}
\def\eg{\emph{e.g}\onedot} 
\def\ie{\emph{i.e}\onedot}

\def\etal{\emph{et al}\onedot}
\def\nextline{\hspace{\textwidth}}

\begin{document}

\title{Adaptive $L_2$ Regularization in \\ Person Re-Identification}

\author{
\IEEEauthorblockN{Xingyang Ni, Liang Fang, Heikki Huttunen}
\IEEEauthorblockA{Faculty of Information Technology and Communication Sciences\\
Tampere University, Finland\\
Email: \{xingyang.ni, liang.fang, heikki.huttunen\}@tuni.fi}
}

\maketitle

\begin{abstract}
We introduce an adaptive $\bm{L_2}$ regularization mechanism in the setting of person re-identification.
In the literature, it is common practice to utilize hand-picked regularization factors which remain constant throughout the training procedure.
Unlike existing approaches, the regularization factors in our proposed method are updated adaptively through backpropagation.
This is achieved by incorporating trainable scalar variables as the regularization factors, which are further fed into a scaled hard sigmoid function.
Extensive experiments on the Market-1501, DukeMTMC-reID and MSMT17 datasets validate the effectiveness of our framework.
Most notably, we obtain state-of-the-art performance on MSMT17, which is the largest dataset for person re-identification.
Source code is publicly available at \url{https://github.com/nixingyang/AdaptiveL2Regularization}.
\end{abstract}

\IEEEpeerreviewmaketitle

\section{Introduction}

Person re-identification involves retrieving corresponding samples from a gallery set based on the appearance of a query sample across multiple cameras.
It is a challenging task since images may differ significantly due to variations in factors such as illumination, camera angle and human pose.
On account of the availability of large-scale datasets~\cite{zheng2015scalable,ristani2016performance,wei2018person}, remarkable progress has been witnessed in recent studies on person re-identification, \eg,
utilizing local feature representations~\cite{varior2016siamese,sun2018beyond},
leveraging extra attribute labels~\cite{su2016deep,lin2019improving},
improving policies for data augmentation~\cite{zhong2017random,dai2019batch},
adding a separate re-ranking step~\cite{zhong2017re,zhou2017efficient}
and switching to video-based datasets~\cite{liu2015spatio,dai2018video}.

$L_2$ regularization imposes constraints on the parameters of neural networks and adds penalties to the objective function during optimization.
It is a commonly adopted technique which can improve the model's generalization ability.
Although some works~\cite{van2017l2,hoffer2018norm,loshchilov2018decoupled,lewkowycz2020training} provide insights on the underlying mechanism of $L_2$ regularization, it is an understudied topic and has not received sufficient attention.
In most literature, $L_2$ regularization is taken for granted, and the text dedicated to it is typically shrunk into one sentence as in~\cite{he2016deep}.
On the other hand, existing approaches assign constant values to regularization factors in the training procedure, and such hyperparameters are hand-picked via hyperparameter optimization which is a tedious and time-consuming process.
The primary purpose of this work is to address the bottleneck of conventional $L_2$ regularization and introduce a mechanism which learns the regularization factors and update those values adaptively.

\newpage
In this paper, our major contributions are twofold:
\begin{itemize}
\item We introduce an adaptive $L_2$ regularization mechanism, which optimizes each regularization factor adaptively as the training procedure progresses.
\item With the proposed framework, we obtain state-of-the-art performance on MSMT17, which is the largest dataset for person re-identification.
\end{itemize}

The rest of this paper is organized as follows.
Section~\ref{section:related_work} reviews important works in person re-identification and $L_2$ regularization.
In Section~\ref{section:proposed_method}, we present the essential components of our baseline, alongside the proposed adaptive $L_2$ regularization mechanism.
Section~\ref{section:experiments} describes the details of our experiments, including datasets, evaluation metrics and comprehensive analysis of our proposed method.
Finally, Section~\ref{section:conclusion} concludes the paper.

\section{Related work}
\label{section:related_work}

In this section, we give a brief overview of two distinct research topics, namely, person re-identification and $L_2$ regularization.

\subsection{Person Re-Identification}

Utilizing local feature representations which are specific to certain regions, has been shown successful.
Varior \etal~\cite{varior2016siamese} propose a Long Short-Term Memory architecture which models the spatial dependency and thus extracts more discriminative local features.
Sun \etal~\cite{sun2018beyond} apply a uniform partition strategy which divides the feature maps evenly into individual parts, and the part-informed features are concatenated to form the final descriptor.

Besides, methods based on auxiliary features are advocated, aiming to utilize extra attributes in addition to the identity labels.
Su \etal~\cite{su2016deep} shows that learning mid-level human attributes can be used to address the challenge of visual appearance variations.
Specifically, an attribute prediction model is trained on an independent dataset which contains the attribute labels.
Lin \etal~\cite{lin2019improving} manually annotate attribute labels which contain detailed local descriptions.
A multi-task network is proposed to learn an embedding for re-identification and also predict the attribute labels.
In addition to the performance improvement in re-identification, such system can speed up the retrieval process by ten times.

\newpage
By applying random manipulations on training samples, data augmentation has played an essential role in suppressing the overfitting issue and improving the generalization of models.
Zhong \etal~\cite{zhong2017random} introduce an approach which erases the pixel values in a random rectangle region during training.
By contrast, Dai \etal~\cite{dai2019batch} suggest dropping the same region for all samples in the same batch.
Such feature dropping branch strengthens the learned features of local regions.

Adding a separate re-ranking step to refine the initial ranking list can lead to significant improvements.
Zhong \etal~\cite{zhong2017re} develop a k-reciprocal encoding method based on the hypothesis that a gallery image is more likely to be a true match if it is similar to the probe in the k-reciprocal nearest neighbours.
Zhou \etal~\cite{zhou2017efficient} rank the predictions with a specified local metric by exploiting negative samples for each online query, rather than implementing a general global metric for all query probes.

Lastly, some works shift the emphasis from image-based to video-based person re-identification.
Liu \etal~\cite{liu2015spatio} introduce a spatio-temporal body-action model which exploits the periodicity exhibited by a walking person in a video sequence.
Alternatively, Dai \etal~\cite{dai2018video} present a learning approach which unifies two modules: one module extracts the features of consecutive frames, and the other module tackles the poor spatial alignment of moving pedestrians.

\subsection{\texorpdfstring{$L_2$}{L2} regularization}

Laarhoven~\cite{van2017l2} prove that $L_2$ regularization would not regularize properly in the presence of normalization operations, \ie, batch normalization~\cite{ioffe2015batch} and weight normalization~\cite{salimans2016weight}.
Instead, $L_2$ regularization will affect the scale of weights, and therefore it has an influence on the effective learning rate.

Similarly, Hoffer \etal~\cite{hoffer2018norm} investigate how does applying weight decay before batch normalization affect learning dynamics.
Combining weight decay and batch normalization would constrain the norm to a small range of values and lead to a more stable step size for the weight direction.
It enables better control over the effective step size through the learning rate.

Later on, Loshchilov \etal~\cite{loshchilov2018decoupled} clarify a long-established misunderstanding that $L_2$ regularization is equivalent to weight decay.
The aforementioned statement does not hold when applying adaptive gradient algorithms, \eg, Adam~\cite{kingma2014adam}.
Furthermore, they suggest decoupling the weight decay from the optimization steps, and it leads to the original formulation of weight decay.

Most recently, Lewkowycz \etal~\cite{lewkowycz2020training} present an empirical study on the relations among the $L_2$ coefficient, the learning rate, and the number of training epochs and the performance of the model.
In a similar manner as learning rate schedules, a manually designed schedule for the L2 parameter is proposed to increase training speed and boost model's performance.

\newpage
\section{Proposed method}
\label{section:proposed_method}

In this section, we first present a minimal setup for person re-identification.
Later on, we explain five components that contribute to significant improvements in performance and use the resulting method as the baseline in our study.
Most importantly, we discuss the proposed adaptive $L_2$ regularization mechanism at the end.

\subsection{Minimal setup}

\begin{figure}[t]
\centering
\includegraphics[width=1.0\linewidth]{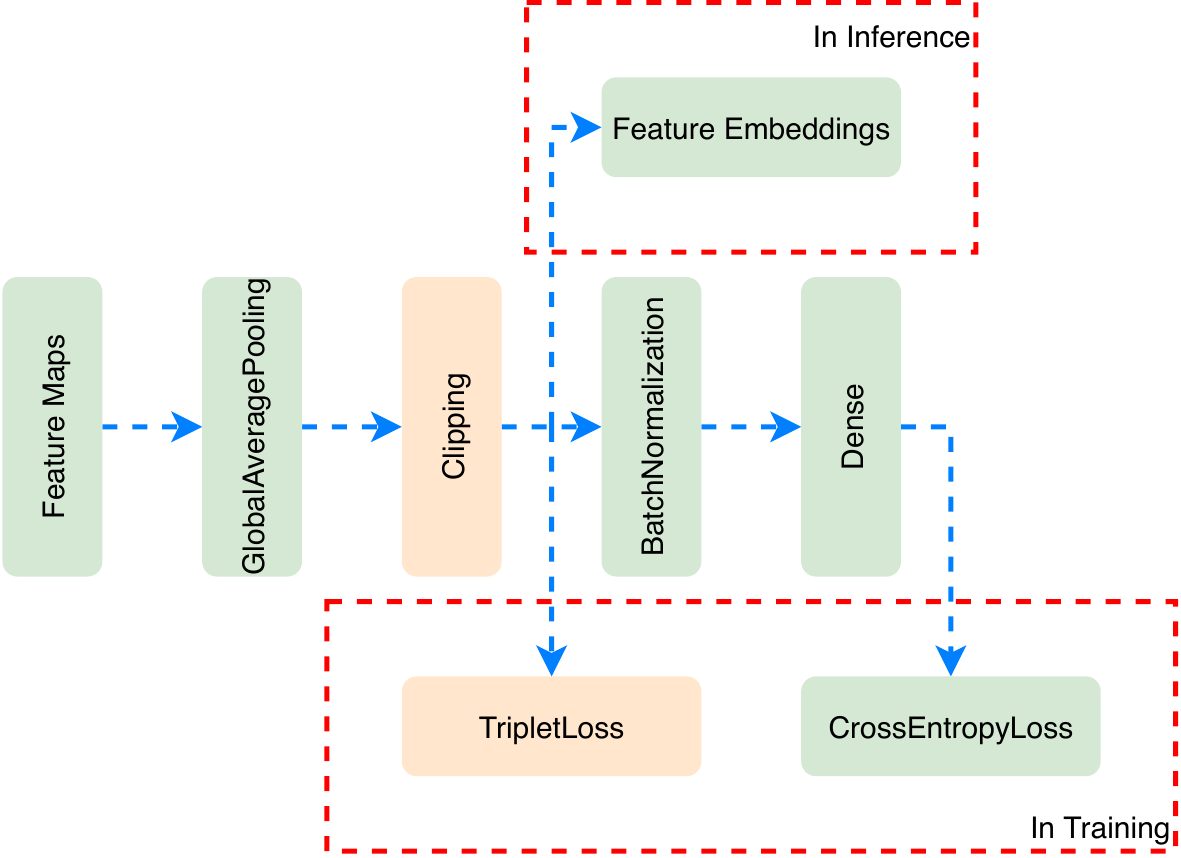}
\caption{
Structure of an objective module.
In the training procedure, two objective functions are applied: triplet loss~\cite{hermans2017defense} and categorical cross-entropy loss.
In the inference procedure, the feature embeddings before the batch normalization~\cite{ioffe2015batch} layer are extracted as the representations.
Note that blocks in yellow are excluded from the minimal setup.
}
\label{figure:objective_module}
\end{figure}

\begin{figure}[t]
\centering
\includegraphics[width=1.0\linewidth]{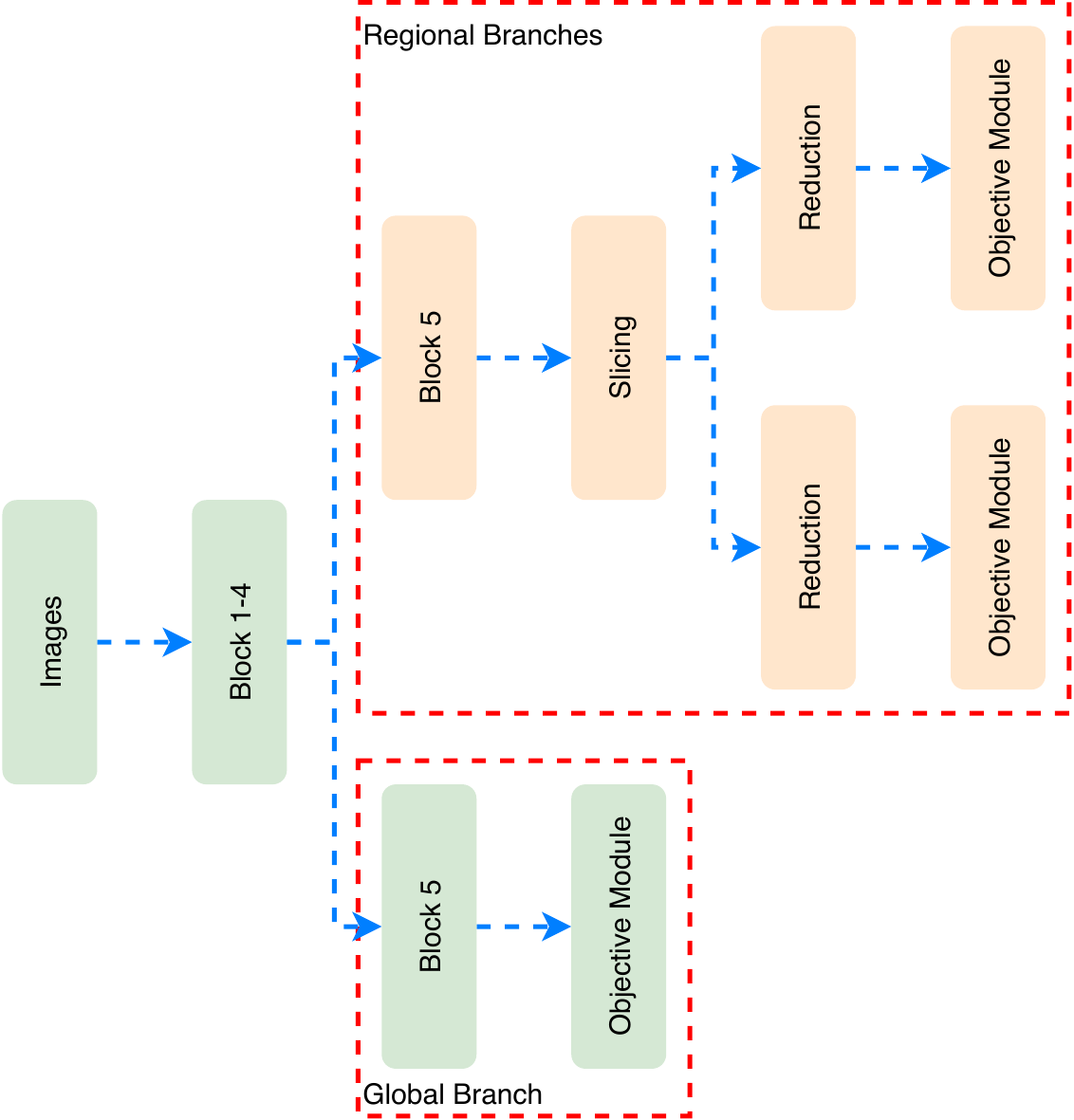}
\caption{
Topology of the overall model.
Feature embeddings from multiple objective modules are concatenated in the inference procedure.
Note that blocks in yellow are excluded from the minimal setup.
}
\label{figure:overall}
\end{figure}

{\bf Backbone:}
ResNet50~\cite{he2016deep}, initialized with ImageNet~\cite{deng2009imagenet} pre-trained weights, is selected as the backbone model.
For convenience, it is separated into five individual blocks, \ie, block 1-5, as illustrated in Figure~\ref{figure:overall}.
Additionally, the stride arguments of the first convolution layer in block 5 are set to 1, rather than default value 2.
This enlarges the feature maps by a scale factor of 2 along with both height and width dimensions while reusing the pre-trained weights and keeping the total amount of parameters identical.

{\bf Objective module:}
Figure~\ref{figure:objective_module} demonstrates the structure of an objective module that converts the feature maps to learning objectives.
A global average pooling layer squeezes the spatial dimensions in the feature maps, and the following batch normalization~\cite{ioffe2015batch} layer generates the normalized feature vectors.
The concluding fully-connected layer does not contain a bias vector, and it produces the predicted probabilities of each unique identity so that the model can be optimized using the categorical cross-entropy loss.
In the inference procedure, the feature embeddings before the batch normalization layer are extracted as the representations, and cosine distance is adopted to measure the distance between two samples.

\newpage
{\bf Overall topology:}
The topology of the overall model is shown in Figure~\ref{figure:overall}.
It is to be observed that the minimal setup only contains the global branch.
Given a batch of images, the individual blocks from the backbone model utilize successively, and an objective module is appended at the end.

{\bf Data augmentation:}
The image is resized to target resolution using a bilinear interpolation method.
Besides, the image is flipped horizontally at random with probability set to 0.5.
Zero paddings are added to all sides of the image, \ie, the top, bottom, left, and right sides.
A random part with target resolution is subsequently cropped.

{\bf Learning rate:}
The learning rate increases linearly from a low value to the pre-defined base learning rate in the early stage of the training procedure, and it is divided by ten once the performance on the validation set plateaus.
On the one hand, the warmup strategy suppresses the distorted gradient issue at the beginning~\cite{liu2019variance}.
On the other hand, periodically reducing the learning rate boosts the performance even further.

{\bf Label smoothing:}
The label smoothing regularization~\cite{szegedy2016rethinking} is applied alongside with the categorical cross-entropy loss function.
Given a sample with ground truth label $y \in \{1,2,\dots,N \}$, the one-hot encoded label $\bm{q}(i)$ equals to $1$ only if the index $i$ is as the same as label $y$, and $0$ otherwise.
The smoothed label introduces a hyperparameter $\epsilon \in (0,1)$ and is calculated as:
\begin{equation}
\bm{q}'(i) = (1 - \epsilon)\bm{q}(i) + \frac{\epsilon}{N}.
\end{equation}

\subsection{Baseline}
\label{section:baseline}

{\bf Triplet loss:}
As highlighted in Figure~\ref{figure:objective_module}, the triplet loss~\cite{hermans2017defense} is applied on the feature embeddings before the batch normalization layer.
It mines the moderate hard triplets instead of all possible combinations of triplets, given that using all possible triplets may destabilize the training procedure.
Considering that multiple loss functions are present, the weighting coefficients of each loss function are set to 1 on account of simplicity.

{\bf Regional branches:}
In addition to the global branch, two regional branches are integrated into the model.
Figure~\ref{figure:overall} illustrates the diagram of those regional branches.
Firstly, the block 5 from the backbone model is replicated, and it is not shared with the global branch.
Secondly, we adopt the uniform partition scheme as in~\cite{sun2018beyond}.
The slicing layer explicitly divides the feature maps into two horizontal stripes.
Lastly, dimensionality reduction is performed using a convolutional layer on each stripe.
Separate objective modules are appended afterwards.
In the inference procedure, feature embeddings from multiple objective modules are concatenated.

{\bf Random erasing:}
In addition to random horizontal flipping, random erasing~\cite{zhong2017random} is utilized in data augmentation.
During training, it erases an area of original images to improve the robustness of the model, especially for the occlusion cases.

{\bf Clipping:}
The clipping layer is inserted between the global average pooling layer and the batch normalization layer in Figure~\ref{figure:objective_module}.
It performs element-wise value clipping so that values in its output are contained in a closed interval.
The clipping layer works in a similar manner as the ReLU-n units~\cite{krizhevsky2010convolutional}, and it relieves optimization difficulties in the succeeding triplet loss~\cite{hermans2017defense}.

{\bf $\bm{L_2}$ regularization:}
Conventional $L_2$ regularization is utilized to all trainable parameters, \ie, the regularization factors remain constant throughout the training procedure.
Additionally, those regularization factors need to be hand-picked via hyperparameter optimization.

\subsection{Adaptive \texorpdfstring{$L_2$}{L2} regularization}

A neural network consists of a set of $N$ distinct parameters,
\begin{equation}
P = \{ \bm{w}_{n} \mid n = 1, \ldots, N \},
\label{equation:distinct_parameters}
\end{equation}
with $P$ containing all trainable parameters.
Each $\bm{w}_{n}$ is an array which could be a vector, a matrix or a 3rd-order tensor.
For example, the kernel and bias terms in a fully-connected layer are a matrix and a vector, respectively.

Conventional $L_2$ regularization imposes an additional penalty term to the objective function, which can be formulated as follows:
\begin{equation}
L_{\lambda}(P) = L(P) + \lambda \sum_{n = 1}^N \|{\bm{w}_{n}}\|_{2}^{2},
\end{equation}
where $L(P)$ and $L_{\lambda}(P)$ denote the original and updated objective functions, respectively.
In our case (see Figures~\ref{figure:objective_module} and~\ref{figure:overall}), $L(P)$ is a weighted sum of triplet loss~\cite{hermans2017defense} and categorical cross-entropy loss functions.
In addition, $\|{\bm{w}_{n}}\|_{2}^{2}$ refers to the square of the $L_2$ norm\footnote{We define $\|{\bm{w}}\|_{2}^{2}$ to denote the sum of squares of all elements also when $\bm{w}$ is a matrix or a 3rd-order tensor.} of $\bm{w}_{n}$, and the constant coefficient $\lambda \in \mathbb{R}_{+}$ defines the regularization strength.

One may wish to add penalties in a different way, \eg, applying lighter regularization in the early layers but stronger in the last ones.
Thus, it is possible to generalize even further, \ie, defining a unique coefficient for each $\|{\bm{w}_{n}}\|_{2}^{2}$:
\begin{equation}
L_{\lambda}(P) = L(P) + \sum_{n = 1}^N \left( \lambda_{n} \|{\bm{w}_{n}}\|_{2}^{2} \right),
\label{equation:l2_regularization}
\end{equation}
where each parameter $\bm{w}_{n}$ is associated with an individual regularization factor $\lambda_{n} \in \mathbb{R}_{+}$.

Obviously, it is infeasible to manually fine-tune those regularization factors $\lambda_{n}$ for $n = 1, \ldots, N$ one by one, since $N$ is in the order of 100 for models trained with ResNet50.
Therefore, we treat them as any other learnable parameters and find suitable values from the data itself.

To make the aforementioned regularization factors adaptive, a straightforward extension is obtained by replacing the pre-defined constant $\lambda_{n}$ with scalar variables which are trainable through backpropagation.
After the modification, Equation~\ref{equation:l2_regularization} remains unchanged while $\lambda_{n} \in \mathbb{R}$.
However, such an approach without any constraints on $\lambda_{n}$ will fail.
Namely, setting negative values for $\lambda_{n}$ allows naively increasing $\|{\bm{w}_{n}}\|_{2}^{2}$ so that $L_{\lambda}(P)$ decreases sharply.
In other words, the $L_2$ regularization penalties would become dominant in the optimization process.
Thus the model collapses and would not learn useful feature embeddings.

To address the collapse problem, we apply the hard sigmoid function which assures that the regularization factor $\lambda_{n}$ would always have non-negative values.
The hard sigmoid function is defined as
\begin{equation}
f(x) =
\begin{cases}
0, & \text{if } x < -c \\
1, & \text{if } x > c \\
x/(2c) + 0.5, & \text{otherwise}. \\
\end{cases}
\end{equation}
In our experiments, we use $c = 2.5$, but any other positive values can be used as well.

The regularization factor $\lambda_{n}$ is obtained by applying the hard sigmoid on the raw parameters as
\begin{equation}
\lambda_{n} = f(\theta_{n}),
\end{equation}
where $\theta_{n} \in \mathbb{R}$ ($n = 1, \ldots, N$) are the trainable scalar variables.
Furthermore, we introduce a hyperparameter $A \in \mathbb{R_{+}}$ which represents the amplitude.
Hence, we get
\begin{equation}
\lambda_{n} = A f(\theta_{n}).
\label{equation:amplitude}
\end{equation}
The amplitude $A$ offers flexibility of avoiding excessively large regularization factors which could deteriorate the training procedure.
Combining Equation~(\ref{equation:l2_regularization}) and~(\ref{equation:amplitude}) gives
\begin{equation}
L_{\lambda}(P) = L(P) + \sum_{n = 1}^N \left( A f(\theta_{n}) \|{\bm{w}_{n}}\|_{2}^{2} \right).
\label{equation:adaptive_l2_regularization}
\end{equation}

\section{Experiments}
\label{section:experiments}

In this section, we explain datasets, evaluation metrics and comprehensive analysis of our proposed method.

\subsection{Datasets}

\begin{table}[t]
\renewcommand{\arraystretch}{1.2}
\caption{
Comparison of three person re-identification datasets, namely,\nextline
Market-1501~\cite{zheng2015scalable}, DukeMTMC-reID~\cite{ristani2016performance} and MSMT17~\cite{wei2018person}.
}
\label{table:comparison_of_datasets}
\centering
\begin{tabular}{@{}lccc@{}}
\toprule
Dataset & Market-1501 & DukeMTMC-reID & MSMT17 \\
\midrule
Train Samples & 12,936 & 16,522 & 32,621 \\
Train Identities & 751 & 702 & 1,041 \\
Test Query Samples & 3,368 & 2,228 & 11,659 \\
Test Gallery Samples & 15,913 & 17,661 & 82,161 \\
Test Identities & 751 & 1,110 & 3,060 \\
Cameras & 6 & 8 & 15 \\
\bottomrule
\end{tabular}
\end{table}

We conduct experiments on three person re-identification datasets, namely, Market-1501~\cite{zheng2015scalable}, DukeMTMC-reID~\cite{ristani2016performance} and MSMT17~\cite{wei2018person}.
Table~\ref{table:comparison_of_datasets} makes a comparison of those datasets.
The MSMT17 dataset outshines the other two due to its large scale.

The Market-1501 dataset is collected with six different cameras in total.
It contains 32,217 images from 1,501 pedestrians, and at least two cameras capture each pedestrian.
The training set includes 751 pedestrians with 12,936 images, while the test set consists of the remaining images from 750 pedestrians and one distractor class.

The DukeMTMC-reID dataset includes 1,404 pedestrians that appear in at least two cameras and 408 pedestrians that appear only in one camera.
The training and test sets contain 16,522 and 19,889 images, respectively.
The query and gallery samples in the test set are randomly split.

The MSMT17 dataset is the largest person re-identification dataset which is publicly available, as of July 2020.
It contains 126,441 images from 4,101 pedestrians, while 3 indoor cameras and 12 outdoor cameras are employed.
In particular, the test set has approximately three times as much samples as the training set.
Such setting motivates the research community to leverage a limited number of training samples that are available since data annotation is costly.

\subsection{Evaluation metrics}

Following the practices in~\cite{zheng2015scalable}, two evaluation metrics are applied to measure the performance, \ie, mean Average Precision (mAP), and Cumulative Matching Characteristic (CMC) rank-k accuracy.
The metrics take the distance matrix between query and gallery samples, in conjunction with the ground truth identities and camera IDs as input arguments.
Gallery samples are discarded if they have been taken from the same camera as the query sample.
As a result, greater emphasis is laid on the performance in the cross-camera setting.

Since the query samples may have multiple ground truth matches in the gallery set, mAP is preferable than rank-k accuracy for the reason that mAP considers both precision and recall.

% \subsection{Implementation details}

% Due to the nature of the triplet loss function~\cite{hermans2017defense}, each batch comprises of samples from both the same and different identities so that the positive and negative exemplars can be generated.
% For each batch, we use images from 16 unique identities, and the number of samples per identity is set to 4.
% All models are trained on images resized to 384x128 for 40,000 iterations using the Adam~\cite{kingma2014adam} optimizer.

% Additionally, we utilize a two-stepped fine-tuning strategy as in~\cite{geng2016deep}.
% In the first step, only the randomly initialized parameters are updated, \ie, layers which are not included in the backbone model.
% In the second step, all parameters are updated instead.
% Such strategy blocks the harmful gradients from those randomly initialized parameters, and it improves the performance on domain adaptation.

\begin{table*}[t]
\renewcommand{\arraystretch}{1.2}
\caption{
Performance comparisons among baseline, adaptive $L_2$ regularization and existing approaches.\nextline
The mAP score on MSMT17 is the most reliable indicator of performance.\nextline
R1: rank-1 accuracy. -: not available. $\dag$: re-ranking~\cite{zhong2017re} is applied.
}
\label{table:performance_comparisons}
\centering
\begin{tabularx}{0.95\textwidth}{@{}lcc*{6}{>{\centering\arraybackslash}X}@{}}
\toprule
\multirow{2}{*}{Method} & \multirow{2}{*}{Venue} & \multirow{2}{*}{Backbone} & \multicolumn{2}{c}{Market-1501} & \multicolumn{2}{c}{DukeMTMC-reID} & \multicolumn{2}{c}{MSMT17} \\
& & & mAP & R1 & mAP & R1 & mAP & R1 \\
\midrule
Annotators~\cite{zhang2017alignedreid} & arXiv 2017 & - & - & 93.5 & - & - & - & - \\
PCB~\cite{sun2018beyond} & ECCV 2018 & ResNet50 & 81.6 & 93.8 & 69.2 & 83.3 & - & - \\
IANet~\cite{hou2019interaction} & CVPR 2019 & ResNet50 & 83.1 & 94.4 & 73.4 & 87.1 & 46.8 & 75.5 \\
AANet~\cite{tay2019aanet} & CVPR 2019 & ResNet50 & 82.5 & 93.9 & 72.6 & 86.4 & - & - \\
CAMA~\cite{yang2019towards} & CVPR 2019 & ResNet50 & 84.5 & 94.7 & 72.9 & 85.8 & - & - \\
DGNet~\cite{zheng2019joint} & CVPR 2019 & ResNet50 & 86.0 & 94.8 & 74.8 & 86.6 & 52.3 & 77.2 \\
OSNet~\cite{zhou2019omni} & ICCV 2019 & OSNet & 84.9 & 94.8 & 73.5 & 88.6 & 52.9 & 78.7 \\
MHN~\cite{chen2019mixed} & ICCV 2019 & ResNet50 & 85.0 & 95.1 & 77.2 & 89.1 & - & - \\
BDB~\cite{dai2019batch} & ICCV 2019 & ResNet50 & 86.7 & 95.3 & 76.0 & 89.0 & - & - \\
BAT-net~\cite{fang2019bilinear} & ICCV 2019 & GoogLeNet & 87.4 & 95.1 & 77.3 & 87.7 & 56.8 & 79.5 \\
SNR~\cite{jin2020style} & CVPR 2020 & ResNet50 & 84.7 & 94.4 & 72.9 & 84.4 & - & - \\
HOReID~\cite{wang2020high} & CVPR 2020 & ResNet50 & 84.9 & 94.2 & 75.6 & 86.9 & - & - \\
RGA-SC~\cite{zhang2020relation} & CVPR 2020 & ResNet50 & 88.4 & 96.1 & - & - & 57.5 & 80.3 \\
SCSN~\cite{chen2020salience} & CVPR 2020 & ResNet50 & 88.5 & 95.7 & 79.0 & 91.0 & 58.5 & 83.8 \\
\midrule
Baseline (Ours) & - & ResNet50 & 87.2\typeout{9503215} & 94.6 & 78.9\typeout{9503216} & 88.0 & 57.7\typeout{9501243} & 79.1 \\
\midrule
\multirow{4}{*}{\shortstack{Adaptive \\ $L_2$ Regularization \\ (Ours)}} & \multirow{4}{*}{-} & ResNet50 & 88.3\typeout{9502037} & 95.3 & 79.9\typeout{9502036} & 88.9 & 59.4\typeout{9502035} & 79.6 \\
& & ResNet101 & 88.6\typeout{9502043} & 94.8 & 80.6\typeout{9502042} & 89.2 & 61.9\typeout{9502040} & 81.3 \\
& & ResNet152 & 88.9\typeout{9502048} & 95.6 & 81.0\typeout{9503472} & 90.2 & 62.2\typeout{9502046} & 81.7 \\
& & ResNet152$^{\dag}$ & 94.4\typeout{9502048} & 96.0 & 90.7\typeout{9503472} & 92.2 & 76.7\typeout{9502046} & 84.9 \\
\bottomrule
\end{tabularx}
\end{table*}

\begin{table}[t]
\renewcommand{\arraystretch}{1.2}
\caption{
Ablation study of baseline using the ResNet50 backbone.\nextline
R1: rank-1 accuracy.
}
\label{table:minimal_setup_vs_baseline}
\centering
\begin{tabular}{@{}lcccccc@{}}
\toprule
\multirow{2}{*}{Method} & \multicolumn{2}{c}{Market-1501} & \multicolumn{2}{c}{DukeMTMC-reID} & \multicolumn{2}{c}{MSMT17} \\
& mAP & R1 & mAP & R1 & mAP & R1 \\
\midrule
Minimal Setup & 28.3\typeout{9503203} & 60.0 & 28.7\typeout{9503204} & 49.9 & 11.4\typeout{9501239} & 34.0 \\
+ Triplet Loss & 79.9\typeout{9503795} & 92.0 & 68.8\typeout{9503208} & 82.2 & 44.0\typeout{9501237} & 70.9 \\
+ Regional Branches & 81.3\typeout{9503209} & 93.3 & 71.2\typeout{9503210} & 84.2 & 47.9\typeout{9501738} & 74.2 \\
+ Random Erasing & 85.8\typeout{9503211} & 94.4 & 76.6\typeout{9503212} & 87.0 & 54.1\typeout{9501739} & 77.0 \\
+ Clipping & 86.8\typeout{9503213} & 94.3 & 78.1\typeout{9503214} & 87.6 & 56.5\typeout{9501627} & 78.4 \\
+ $L_2$ regularization & 87.2\typeout{9503215} & 94.6 & 78.9\typeout{9503216} & 88.0 & 57.7\typeout{9501243} & 79.1 \\
\bottomrule
\end{tabular}
\end{table}

\subsection{Ablation study of baseline}

The baseline differs from the minimal setup in five aspects, as discussed in Section~\ref{section:baseline}.
Table~\ref{table:minimal_setup_vs_baseline} presents an ablation study to demonstrate how each component contributes to the performance on person re-identification.
On the one hand, the triplet loss~\cite{hermans2017defense} brings the most significant improvements on all three datasets.
The boost is due to the fact that the triplet loss is applied to the feature embeddings which are retrieved in the inference procedure (see Figure~\ref{figure:objective_module}).
Since the triplet loss directly optimizes the model in a manner comparable to similarity search, it closes the gap between the training and inference procedures.
On the other hand, the other four components bring moderate improvements.
It is conceivable that the model reaches better generalization by using hand-picked $L_2$ regularization factors which remain constant throughout the training procedure.

\subsection{Comparisons with existing approaches}

Table~\ref{table:performance_comparisons} shows performance comparisons among baseline, adaptive $L_2$ regularization and existing approaches.

Firstly, all methods listed in Table~\ref{table:performance_comparisons} have surpassed the best-performing human annotators~\cite{zhang2017alignedreid} on the Market-1501 dataset.
In light of the scale of the Market-1501 and DukeMTMC-reID datasets (see Table~\ref{table:comparison_of_datasets}), these two small-scale datasets might have been saturated and more emphasis should be put on the MSMT17 dataset.
Since mAP is preferable than rank-k accuracy, the mAP score on MSMT17 is the most reliable indicator of performance.

Secondly, the proposed adaptive $L_2$ regularization mechanism contributes with decent improvements to the baseline, especially on MSMT17 in which the mAP score increases from $57.7\%$ to $59.4\%$.
Among methods which utilize the ResNet50 backbone, it is noteworthy that the adaptive $L_2$ regularization method obtains the state-of-the-art performance on DukeMTMC-reID and MSMT17, very close to state-of-the-art performance on Market-1501.

Last but not least, deeper backbones (\ie, ResNet101 and ResNet152) further improve the performance, at the cost of extra computations.
Attributed to the re-ranking~\cite{zhong2017re} method which exploits the test data in the inference procedure, new milestones have been accomplished, \ie, the mAP scores on Market-1501, DukeMTMC-reID and MSMT17 stand at $94.4\%$, $90.7\%$ and $76.7\%$, respectively.

\subsection{Quantitative analysis of regularization factors}

\begin{figure}[t]
\centering
\includegraphics[width=1.0\linewidth]{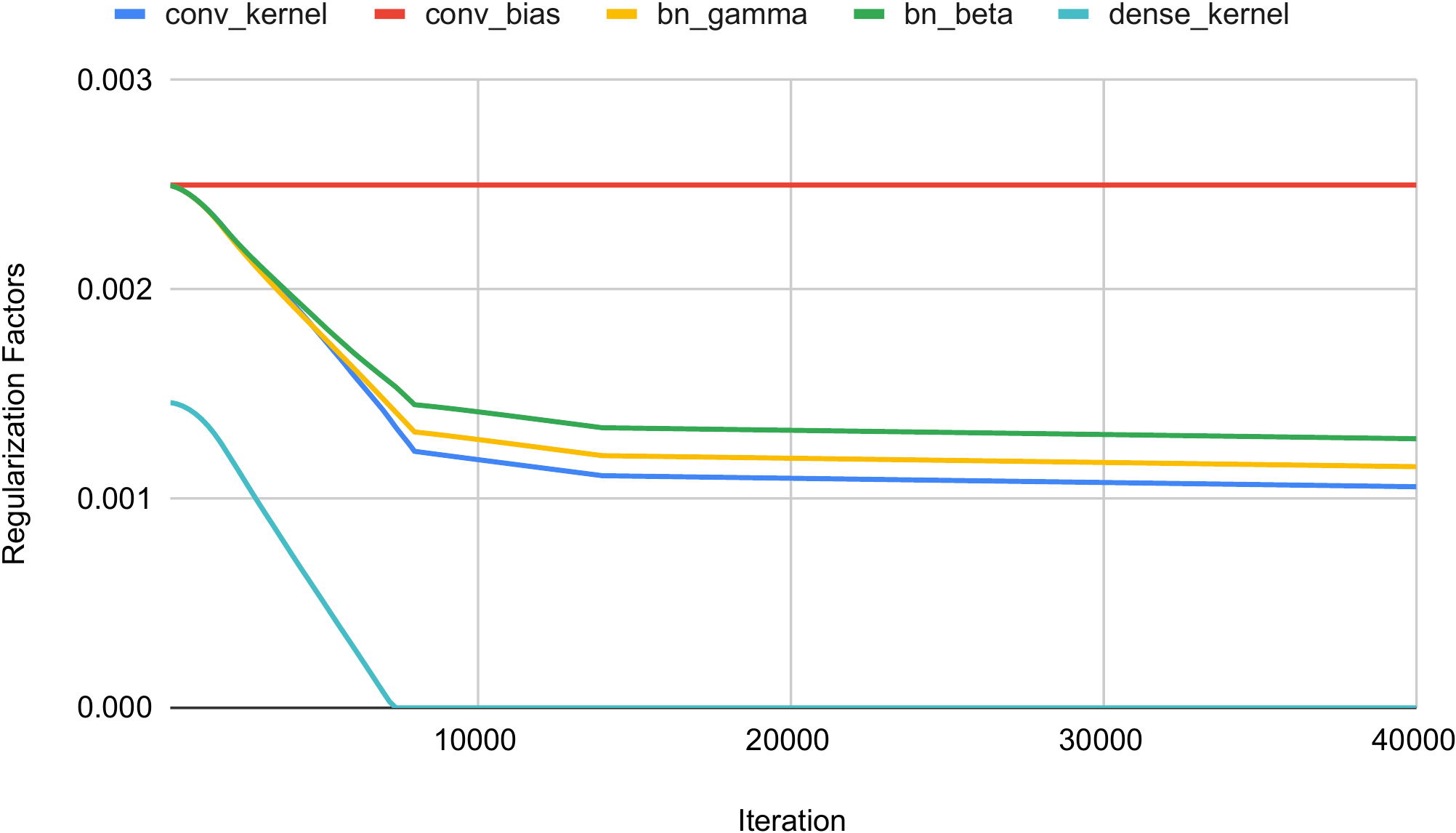}
\caption{
The median value of regularization factors in each category, with respect to the number of iterations.
}
\label{figure:median_vs_epoch}
\end{figure}

Depending on the associated distinct parameter (see Equation~\ref{equation:distinct_parameters}), the regularization factors can be classified into five categories:
\textit{conv\_kernel}, \textit{conv\_bias}, \textit{bn\_gamma}, \textit{bn\_beta} and \textit{dense\_kernel},
where \textit{conv}, \textit{bn} and \textit{dense} denote the convolutional, batch normalization and fully-connected layers.
In the following, we examine the regularization factors for a model trained on MSMT17 using the ResNet50 backbone.

Figure~\ref{figure:median_vs_epoch} visualizes the median value of regularization factors in each category, with respect to the number of iterations.
Note that the learning rate gets reduced at iterations $8000$ and $14000$.
While \textit{conv\_kernel}, \textit{bn\_gamma} and \textit{bn\_beta} behave similarly, \textit{conv\_bias} remains constant throughout the training procedure and \textit{dense\_kernel} drops to $0$ in the early stage.

\begin{figure}[t]
\centering
\includegraphics[width=1.0\linewidth]{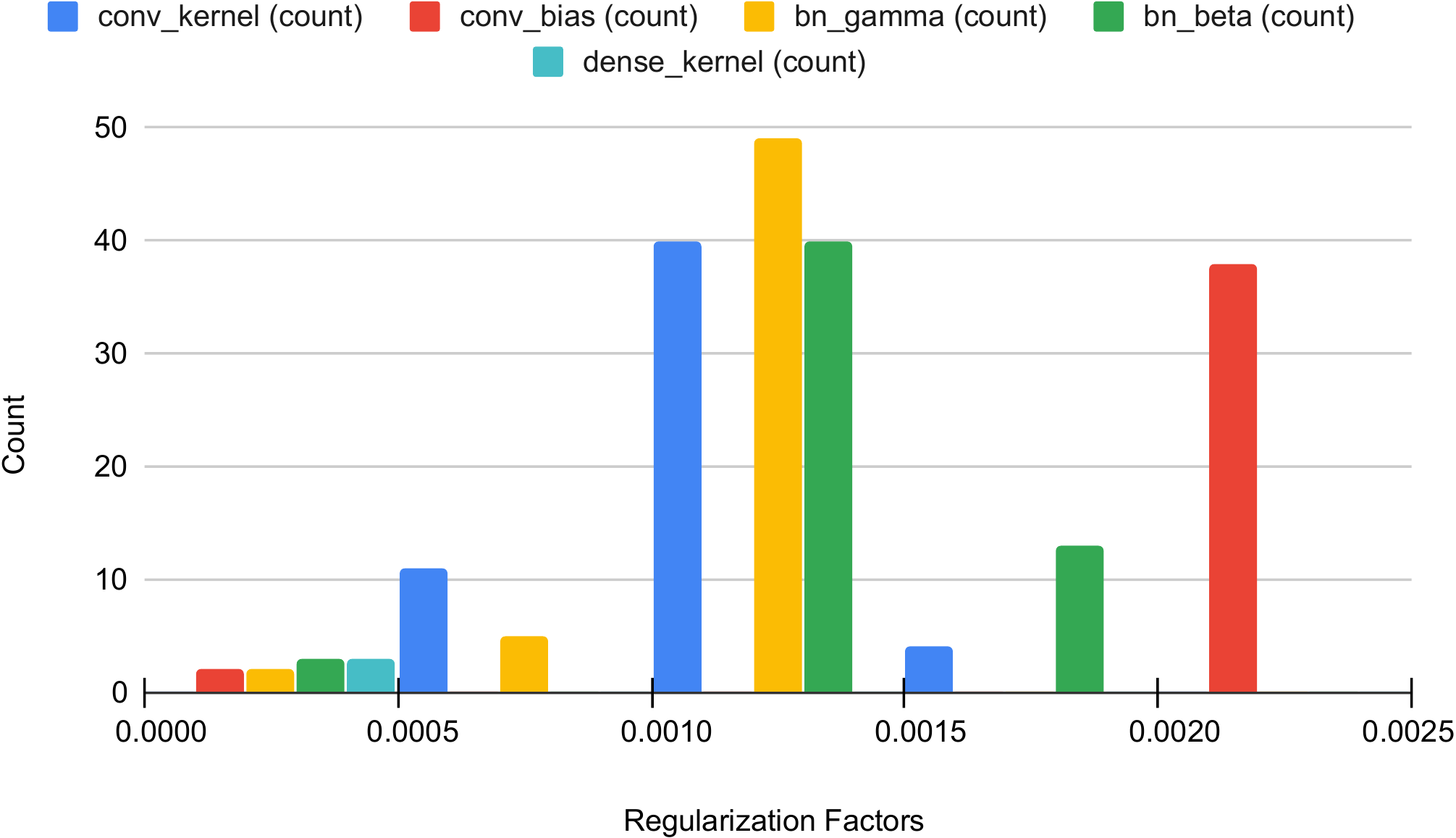}
\caption{
Histogram of regularization factors in the last epoch.
}
\label{figure:last_epoch}
\end{figure}

Figure~\ref{figure:last_epoch} demonstrates a histogram of regularization factors in the last epoch, \ie, the training procedure completes.
The interval $[0, 0.0025]$ is divided evenly into five buckets.
For regularization factors from the same category, the values could differ significantly, \eg,
$2$ and $38$ regularization factors from \textit{conv\_bias} fall within the interval $[0, 0.0005]$ and $[0.0020, 0.0025]$, respectively.
To be specific, the regularization factors from \textit{conv\_bias} in the two \textit{Reduction} blocks are $0$ (see Figure~\ref{figure:overall}).
If omitting the effects of the \textit{Clipping} layer in Figure~\ref{figure:objective_module}, those convolutional layers are followed by batch normalization layers which intrinsically cancels out the bias terms in aforementioned convolutional layers.
Consequently, such regularization factors would converge to $0$.
In summary, this phenomenon reflects the superiority of our proposed method, in which each regularization factor is optimized separately.

\subsection{Qualitative analysis of predictions}

\begin{figure}[t]
\centering
\includegraphics[width=1.0\linewidth]{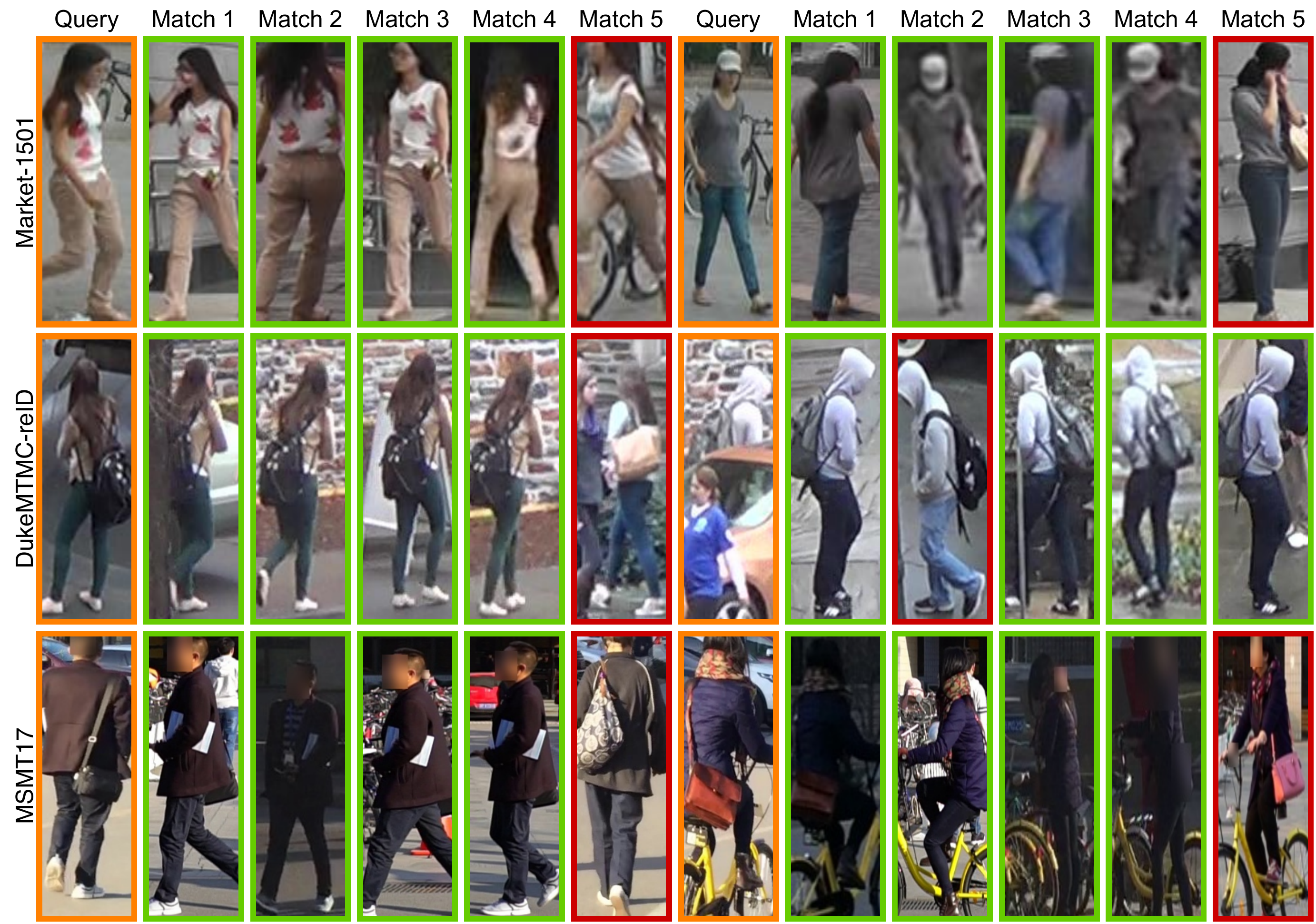}
\caption{
Selected query samples with corresponding top 5 matches from the gallery set.
Images with orange, green and red border are query samples, correct matches and erroneous matches, respectively.
}
\label{figure:prediction}
\end{figure}

Figure~\ref{figure:prediction} illustrates selected query samples with corresponding top 5 matches from the gallery set.
Although query samples and erroneous matches may have similar appearances, minor differences could be observed under careful inspection, \eg, the dissimilarity between backpacks.
Furthermore, our models could retrieve correct matches even in the presence of large illumination changes, \eg, the two examples from the MSMT17 dataset.

\section{Conclusion}
\label{section:conclusion}

In this work, we revisited $L_2$ regularization in neural networks.
Unlike employing constant and hand-picked regularization factors, our proposed method optimizes the strength of $L_2$ regularization adaptively through backpropagation.
More specifically, we applied a scaled hard sigmoid function to trainable scalar variables and used those as the regularization factors.
Extensive experiments substantiate that our framework boosts model's performance, and we obtained state-of-the-art performance on MSMT17 which is the largest person re-identification dataset.
Although the current study is constrained within the domain of person re-identification, it is self-evident that the proposed adaptive $L_2$ regularization mechanism can be seamlessly integrated into the training procedure of any neural networks.
Accordingly, further studies can be conducted on other topics, \eg, image classification, image retrieval and object detection.

\section*{Acknowledgement}
This work was financially supported by Business Finland project 408/31/2018 MIDAS.

\newpage
\bibliographystyle{IEEEtran}
\bibliography{nixingyang_references}

% Generated by IEEEtran.bst, version: 1.12 (2007/01/11)
\begin{thebibliography}{10}
\providecommand{\url}[1]{#1}
\csname url@samestyle\endcsname
\providecommand{\newblock}{\relax}
\providecommand{\bibinfo}[2]{#2}
\providecommand{\BIBentrySTDinterwordspacing}{\spaceskip=0pt\relax}
\providecommand{\BIBentryALTinterwordstretchfactor}{4}
\providecommand{\BIBentryALTinterwordspacing}{\spaceskip=\fontdimen2\font plus
\BIBentryALTinterwordstretchfactor\fontdimen3\font minus
  \fontdimen4\font\relax}
\providecommand{\BIBforeignlanguage}[2]{{%
\expandafter\ifx\csname l@#1\endcsname\relax
\typeout{** WARNING: IEEEtran.bst: No hyphenation pattern has been}%
\typeout{** loaded for the language `#1'. Using the pattern for}%
\typeout{** the default language instead.}%
\else
\language=\csname l@#1\endcsname
\fi
#2}}
\providecommand{\BIBdecl}{\relax}
\BIBdecl

\bibitem{zheng2015scalable}
L.~Zheng, L.~Shen, L.~Tian, S.~Wang, J.~Wang, and Q.~Tian, ``{Scalable person
  re-identification: A benchmark},'' in \emph{Proceedings of the IEEE
  International Conference on Computer Vision}, 2015, pp. 1116--1124.

\bibitem{ristani2016performance}
E.~Ristani, F.~Solera, R.~Zou, R.~Cucchiara, and C.~Tomasi, ``{Performance
  measures and a data set for multi-target, multi-camera tracking},'' in
  \emph{European Conference on Computer Vision}.\hskip 1em plus 0.5em minus
  0.4em\relax Springer, 2016, pp. 17--35.

\bibitem{wei2018person}
L.~Wei, S.~Zhang, W.~Gao, and Q.~Tian, ``{Person transfer gan to bridge domain
  gap for person re-identification},'' in \emph{Proceedings of the IEEE
  Conference on Computer Vision and Pattern Recognition}, 2018, pp. 79--88.

\bibitem{varior2016siamese}
R.~R. Varior, B.~Shuai, J.~Lu, D.~Xu, and G.~Wang, ``{A siamese long short-term
  memory architecture for human re-identification},'' in \emph{European
  conference on computer vision}.\hskip 1em plus 0.5em minus 0.4em\relax
  Springer, 2016, pp. 135--153.

\bibitem{sun2018beyond}
Y.~Sun, L.~Zheng, Y.~Yang, Q.~Tian, and S.~Wang, ``{Beyond part models: Person
  retrieval with refined part pooling (and a strong convolutional baseline)},''
  in \emph{Proceedings of the European Conference on Computer Vision}, 2018,
  pp. 480--496.

\bibitem{su2016deep}
C.~Su, S.~Zhang, J.~Xing, W.~Gao, and Q.~Tian, ``{Deep attributes driven
  multi-camera person re-identification},'' in \emph{European conference on
  computer vision}.\hskip 1em plus 0.5em minus 0.4em\relax Springer, 2016, pp.
  475--491.

\bibitem{lin2019improving}
Y.~Lin, L.~Zheng, Z.~Zheng, Y.~Wu, Z.~Hu, C.~Yan, and Y.~Yang, ``{Improving
  person re-identification by attribute and identity learning},'' \emph{Pattern
  Recognition}, 2019.

\bibitem{zhong2017random}
Z.~Zhong, L.~Zheng, G.~Kang, S.~Li, and Y.~Yang, ``{Random erasing data
  augmentation},'' \emph{arXiv preprint arXiv:1708.04896}, 2017.

\bibitem{dai2019batch}
Z.~Dai, M.~Chen, X.~Gu, S.~Zhu, and P.~Tan, ``{Batch DropBlock network for
  person re-identification and beyond},'' in \emph{Proceedings of the IEEE
  International Conference on Computer Vision}, 2019, pp. 3691--3701.

\bibitem{zhong2017re}
Z.~Zhong, L.~Zheng, D.~Cao, and S.~Li, ``{Re-ranking person re-identification
  with k-reciprocal encoding},'' in \emph{Proceedings of the IEEE Conference on
  Computer Vision and Pattern Recognition}, 2017, pp. 1318--1327.

\bibitem{zhou2017efficient}
J.~Zhou, P.~Yu, W.~Tang, and Y.~Wu, ``{Efficient online local metric adaptation
  via negative samples for person re-identification},'' in \emph{Proceedings of
  the IEEE International Conference on Computer Vision}, 2017, pp. 2420--2428.

\bibitem{liu2015spatio}
K.~Liu, B.~Ma, W.~Zhang, and R.~Huang, ``{A spatio-temporal appearance
  representation for viceo-based pedestrian re-identification},'' in
  \emph{Proceedings of the IEEE International Conference on Computer Vision},
  2015, pp. 3810--3818.

\bibitem{dai2018video}
J.~Dai, P.~Zhang, D.~Wang, H.~Lu, and H.~Wang, ``{Video person
  re-identification by temporal residual learning},'' \emph{IEEE Transactions
  on Image Processing}, vol.~28, no.~3, pp. 1366--1377, 2018.

\bibitem{van2017l2}
T.~Van~Laarhoven, ``{L2 regularization versus batch and weight
  normalization},'' \emph{arXiv preprint arXiv:1706.05350}, 2017.

\bibitem{hoffer2018norm}
E.~Hoffer, R.~Banner, I.~Golan, and D.~Soudry, ``{Norm matters: efficient and
  accurate normalization schemes in deep networks},'' in \emph{Advances in
  Neural Information Processing Systems}, 2018, pp. 2160--2170.

\bibitem{loshchilov2018decoupled}
I.~Loshchilov and F.~Hutter, ``{Decoupled weight decay regularization},''
  \emph{arXiv preprint arXiv:1711.05101}, 2018.

\bibitem{lewkowycz2020training}
A.~Lewkowycz and G.~Gur-Ari, ``{On the training dynamics of deep networks with
  {\$} L{\_}2 {\$} regularization},'' \emph{arXiv preprint arXiv:2006.08643},
  2020.

\bibitem{he2016deep}
K.~He, X.~Zhang, S.~Ren, and J.~Sun, ``{Deep residual learning for image
  recognition},'' in \emph{Proceedings of the IEEE Conference on Computer
  Vision and Pattern Recognition}, 2016, pp. 770--778.

\bibitem{ioffe2015batch}
S.~Ioffe and C.~Szegedy, ``{Batch normalization: Accelerating deep network
  training by reducing internal covariate shift},'' \emph{arXiv preprint
  arXiv:1502.03167}, 2015.

\bibitem{salimans2016weight}
T.~Salimans and D.~P. Kingma, ``{Weight normalization: A simple
  reparameterization to accelerate training of deep neural networks},'' in
  \emph{Advances in neural information processing systems}, 2016, pp. 901--909.

\bibitem{kingma2014adam}
D.~P. Kingma and J.~Ba, ``{Adam: A method for stochastic optimization},''
  \emph{arXiv preprint arXiv:1412.6980}, 2014.

\bibitem{hermans2017defense}
A.~Hermans, L.~Beyer, and B.~Leibe, ``{In defense of the triplet loss for
  person re-identification},'' \emph{arXiv preprint arXiv:1703.07737}, 2017.

\bibitem{deng2009imagenet}
J.~Deng, W.~Dong, R.~Socher, L.-J. Li, K.~Li, and L.~Fei-Fei, ``{ImageNet: A
  large-scale hierarchical image database},'' in \emph{IEEE Conference on
  Computer Vision and Pattern Recognition}.\hskip 1em plus 0.5em minus
  0.4em\relax Ieee, 2009, pp. 248--255.

\bibitem{liu2019variance}
L.~Liu, H.~Jiang, P.~He, W.~Chen, X.~Liu, J.~Gao, and J.~Han, ``{On the
  variance of the adaptive learning rate and beyond},'' \emph{arXiv preprint
  arXiv:1908.03265}, 2019.

\bibitem{szegedy2016rethinking}
C.~Szegedy, V.~Vanhoucke, S.~Ioffe, J.~Shlens, and Z.~Wojna, ``{Rethinking the
  inception architecture for computer vision},'' in \emph{Proceedings of the
  IEEE Conference on Computer Vision and Pattern Recognition}, 2016, pp.
  2818--2826.

\bibitem{krizhevsky2010convolutional}
A.~Krizhevsky and G.~Hinton, ``{Convolutional deep belief networks on
  cifar-10},'' \emph{Unpublished manuscript}, vol.~40, no.~7, pp. 1--9, 2010.

\bibitem{zhang2017alignedreid}
X.~Zhang, H.~Luo, X.~Fan, W.~Xiang, Y.~Sun, Q.~Xiao, W.~Jiang, C.~Zhang, and
  J.~Sun, ``{Alignedreid: Surpassing human-level performance in person
  re-identification},'' \emph{arXiv preprint arXiv:1711.08184}, 2017.

\bibitem{hou2019interaction}
R.~Hou, B.~Ma, H.~Chang, X.~Gu, S.~Shan, and X.~Chen,
  ``{Interaction-and-aggregation network for person re-identification},'' in
  \emph{Proceedings of the IEEE Conference on Computer Vision and Pattern
  Recognition}, 2019, pp. 9317--9326.

\bibitem{tay2019aanet}
C.-P. Tay, S.~Roy, and K.-H. Yap, ``{Aanet: Attribute attention network for
  person re-identifications},'' in \emph{Proceedings of the IEEE Conference on
  Computer Vision and Pattern Recognition}, 2019, pp. 7134--7143.

\bibitem{yang2019towards}
W.~Yang, H.~Huang, Z.~Zhang, X.~Chen, K.~Huang, and S.~Zhang, ``{Towards rich
  feature discovery with class activation maps augmentation for person
  re-identification},'' in \emph{Proceedings of the IEEE Conference on Computer
  Vision and Pattern Recognition}, 2019, pp. 1389--1398.

\bibitem{zheng2019joint}
Z.~Zheng, X.~Yang, Z.~Yu, L.~Zheng, Y.~Yang, and J.~Kautz, ``{Joint
  discriminative and generative learning for person re-identification},'' in
  \emph{Proceedings of the IEEE Conference on Computer Vision and Pattern
  Recognition}, 2019, pp. 2138--2147.

\bibitem{zhou2019omni}
K.~Zhou, Y.~Yang, A.~Cavallaro, and T.~Xiang, ``{Omni-scale feature learning
  for person re-identification},'' in \emph{Proceedings of the IEEE
  International Conference on Computer Vision}, 2019, pp. 3702--3712.

\bibitem{chen2019mixed}
B.~Chen, W.~Deng, and J.~Hu, ``{Mixed high-order attention network for person
  re-identification},'' in \emph{Proceedings of the IEEE International
  Conference on Computer Vision}, 2019, pp. 371--381.

\bibitem{fang2019bilinear}
P.~Fang, J.~Zhou, S.~K. Roy, L.~Petersson, and M.~Harandi, ``{Bilinear
  attention networks for person retrieval},'' in \emph{Proceedings of the IEEE
  International Conference on Computer Vision}, 2019, pp. 8030--8039.

\bibitem{jin2020style}
X.~Jin, C.~Lan, W.~Zeng, Z.~Chen, and L.~Zhang, ``{Style normalization and
  restitution for generalizable person re-identification},'' in
  \emph{Proceedings of the IEEE/CVF Conference on Computer Vision and Pattern
  Recognition}, 2020, pp. 3143--3152.

\bibitem{wang2020high}
G.~Wang, S.~Yang, H.~Liu, Z.~Wang, Y.~Yang, S.~Wang, G.~Yu, E.~Zhou, and
  J.~Sun, ``{High-Order Information Matters: Learning Relation and Topology for
  Occluded Person Re-Identification},'' in \emph{Proceedings of the IEEE/CVF
  Conference on Computer Vision and Pattern Recognition}, 2020, pp. 6449--6458.

\bibitem{zhang2020relation}
Z.~Zhang, C.~Lan, W.~Zeng, X.~Jin, and Z.~Chen, ``{Relation-Aware Global
  Attention for Person Re-identification},'' in \emph{Proceedings of the
  IEEE/CVF Conference on Computer Vision and Pattern Recognition}, 2020, pp.
  3186--3195.

\bibitem{chen2020salience}
X.~Chen, C.~Fu, Y.~Zhao, F.~Zheng, J.~Song, R.~Ji, and Y.~Yang,
  ``{Salience-Guided Cascaded Suppression Network for Person
  Re-Identification},'' in \emph{Proceedings of the IEEE/CVF Conference on
  Computer Vision and Pattern Recognition}, 2020, pp. 3300--3310.

\end{thebibliography}

\end{document}